\pdfoutput=1

\documentclass[11pt]{article}

\usepackage{acl}

\usepackage{times}
\usepackage{latexsym}
\usepackage{graphicx}
\usepackage{color,soul}
\usepackage{hyperref}
\usepackage{natbib}

\DeclareMathAlphabet      {\mathbfit}{OML}{cmm}{b}{it}

\usepackage[T1]{fontenc}

\usepackage[utf8]{inputenc}

\usepackage{microtype}

\usepackage{amsmath}
\usepackage{amssymb}

\usepackage{caption}
\usepackage{subcaption}

\usepackage{booktabs}
\usepackage{multirow}

\title{Using Multi-Encoder Fusion Strategies to Improve Personalized Response Selection}

\author{Souvik Das \space Sougata Saha \space Rohini K. Srihari \\
        \texttt{\{souvikda, sougatas, rohini\}@buffalo.edu} \\
        Department of Computer Science and Engineering, University at Buffalo, NY. \\}

\begin{document}
\maketitle
\begin{abstract}
Personalized response selection systems are generally grounded on \textit{persona}. However, a correlation exists between persona and empathy, which these systems do not explore well. Also, when a contradictory or off-topic response is selected, \textit{faithfulness} to the conversation context plunges. This paper attempts to address these issues by proposing a suite of fusion strategies that capture the interaction between persona, emotion, and entailment information of the utterances. Ablation studies on \texttt{Persona-Chat} dataset show that incorporating emotion and entailment improves the accuracy of response selection. We combine our fusion strategies and concept-flow encoding to train a BERT-based model which outperforms the previous methods by margins larger than 2.3\% on original personas and 1.9\% on revised personas in terms of \textbf{hits@1} (top-1 accuracy), achieving a new state-of-the-art performance on the \texttt{Persona-Chat} dataset.
\end{abstract}

\section{Introduction}

\begin{figure}[h]
\centering
\includegraphics[width=0.5\textwidth]{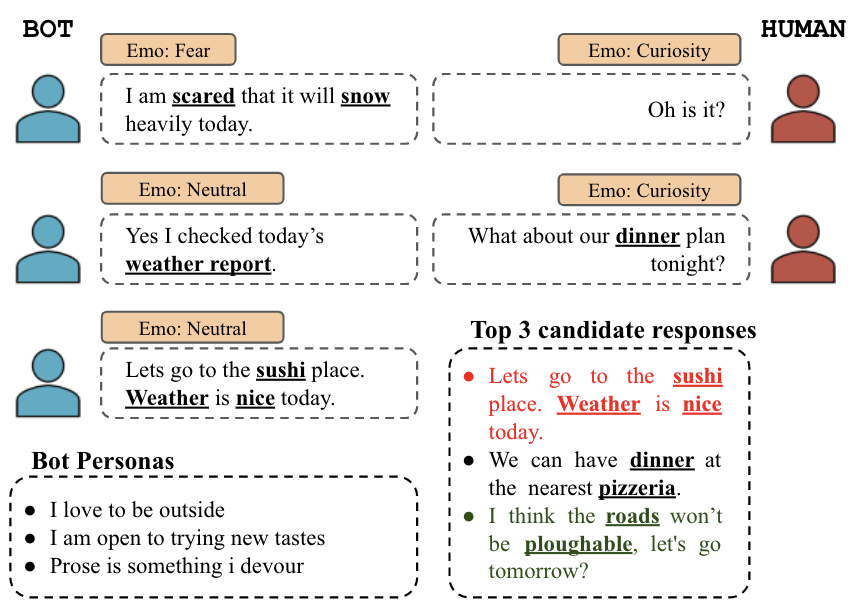}
\caption{For this conversation, the selected candidate response directly contradicts the context. Also, the bot’s persona influences the response selection, while the situational emotions and concepts get ignored. The underlines phrases/words denote the concepts.}
\label{fig:example}
\end{figure}

Currently, most response selection systems tend to perform well in most cases \cite{ir1, ir2, ir3, gu2020speakeraware}. However, these re-ranking systems have the poor capability to detect and evade contradictory responses. Responses selected by these systems often contradict previous utterances, and any form of contradiction may disrupt the flow of conversation. Previous research has attempted to incorporate persona while selecting \cite{partner, partner2} or generating \cite{wu-etal-2021-personalized} responses to maintain consistency. Additionally, a correlation exists between persona with personality \cite{p2p} , which influences empathy \cite{richendoller1994exploring}. \citeauthor{pec} presented a multi-domain dataset collected from several empathetic Reddit threads contributing towards persona-based empathetic conversations. Nevertheless, no one has studied the emotion-persona interplay in data presented in a more natural form. Figure \ref{fig:example} depicts situational emotion sometimes needs more preference than the chatbot's persona in response selection.

On the contrary, different personality traits are related to emotion regulation difficulties \cite{POLLOCK2016168}. Due to this, a person's expected emotions can deviate based on his persona. Besides that, we also observe that concepts discussed in a conversational flow play an important role in response selection. However, no one has incorporated this in response selection.

\begin{table}[h]
\resizebox{\columnwidth}{!}{
    \begin{tabular}{@{}llll@{}}
    \toprule
    Model    & Emotion Inappropriate(\%) & Contradictory(\%)             & Off-topic(\%)                \\ \midrule
    BERT-CRA & \multicolumn{1}{r}{7.35} & \multicolumn{1}{r}{11.88} & \multicolumn{1}{r}{12.3} \\ \bottomrule
    \end{tabular}
}
\caption[Statistics of issues reported in the test split of \texttt{Persona-Chat} inferred by BERT-CRA]{Statistics of issues reported in the test split of \texttt{Persona-Chat} inferred by BERT-CRA \cite{partner} \protect\footnotemark}

\label{tab1}
\end{table}

\footnotetext{Insights drawn from the human evaluation done on 500 randomly selected data-points from self-persona original and partner-persona original sets of \texttt{Persona-Chat}}

We can infer the significance of these problems from Table \ref{tab1}. So, to increase the usability of the personalized response selection systems, all these fundamental problems need to be addressed. We automatically annotate \texttt{Persona-Chat} \cite{zhang2018personalizing} dataset using a series of classifiers and rule-based modules. We model emotion-persona interaction, context-response entailment, and concept-flow using the annotations. To compare the ability of annotated features to enhance the emotion-persona interaction, contradiction avoidance, and adherence to the concept flow, we perform preliminary experiments by devising independent encoders based on BERT. Our baseline model extends BERT-CRA \cite{partner} where we introduce an additional bot-encoder to represent the bot-utterances better. Subsequently, we propose three fusion strategies, emotion-aware(EmA), entailment-aware(EnA), persona-entailment-aware(P-EmA). These fusion strategies are designed based on emotion-persona interaction or persona-entailment information. We propose a concept-flow encoding technique that matches relevant concepts from the context and candidate responses with these fusion strategies.


We test our proposed methods on the \texttt{Persona-Chat} dataset with our automatic annotation. The results show that a model trained on a combination of our proposed fusion strategies outperforms the current state-of-the-art model by a margin of 2.3\% in terms of  top-1 accuracy \textbf{hits@1}. 

In summary, the contributions of this paper are three-fold. (1)Automatically annotate \texttt{Persona-Chat} dataset with utterance level emotion, entailment, and concept information to provide extra supervision. (2) A suite of fusion strategies and a concept-flow encoder which are designed and implemented into a series of models, aiming to explore the impact of emotion, entailment, and concept-flow in the task of response selection. (3)  Experimental results demonstrate that our proposed models outperform the existing state-of-the-art models by significant margins on the widely used \texttt{Persona-Chat} response selection benchmark. 

\section{Related Works}

\subsection{Personalized Response Selection}
Chit-chat models typically trained over many dialogues with different speakers lack a consistent personality and explicit long-term memory. These models produce an utterance given only a recent dialogue history. \citeauthor{li-etal-2016-persona}  proposed a persona-based neural conversation model to capture individual characteristics such as background information and speaking style. \cite{zhang2018personalizing} has constructed \texttt{Persona-Chat} dataset to build personalized dialog systems; this is by far the largest public dataset containing million-turn dialog conditioned on persona. Many benchmarks have been established for this dataset. For example, \cite{mazare-etal-2018-training} proposed the fine-tuned Persona-Chat (FT-PC) model, which first pre-trained models using a large-scale corpus based on Reddit to extract valuable dialogues conditioned on personas and then fine-tuned these pre-trained models on the \texttt{Persona-Chat} dataset. \cite{wolf2019transfertransfo, liu2020impress} also employed the pre-trained language model(GPT)  for building personalized dialogue agents. \cite{gu-etal-2020-filtering} proposed filtering before iteratively referring (FIRE) to ground the conversation on the given knowledge and then perform the deep and iterative matching. \cite{partner} explored a new direction by proposing four persona fusion strategies, thereby incorporating partner persona in response selection.
\subsection{Faithfulness to Context}
Faithfulness in conversational systems to conversation context or knowledge is a very broad topic that can range from decreasing fact hallucination\cite{chen-etal-2021-improving}, reducing contradictory responses, staying on topic, etc. \cite{rashkin2021increasing} has used additional inputs to act as stylistic controls that encourage the model to generate responses that are faithful to a provided evidence or knowledge. However, no one has studied the level of faithfulness the current personalized response selection systems exhibit to the conversation history. Thus, this paper thoroughly explores the impact of utilizing utterance-level emotions, entailment, and concepts on the performance of personalized response selection. 
\section{Dataset} \label{dataset}
In this work, we extend \texttt{Persona-Chat} \cite{zhang2018personalizing} and augment it with a series of annotators.  The dataset consists of 8939 complete dialogues for training, 1000 for validation, and 968 for testing. Responses are selected at every turn of a conversation sequence, resulting in 65719 context-responses pairs for training, 7801 for validation, and 7512 for testing. The positive and negative response ratio is 1:19 in the training, validation, and testing sets. There are 955 possible personas for training, 100 for validation, and 100 for testing, consisting of 3 to 5 profile sentences. A revised version of persona descriptions is also provided by rephrasing, generalizing, or specializing the original ones to make this task more challenging.
\section{Automatic Dataset Annotation}
We have annotated the \texttt{Persona-Chat} with the help of a series of automatic annotation schemes. Since we are studying the effect of emotions in personalized response selection, we assign emotion labels to the personas, context-utterances, and candidate responses using an emotion classifier. Personas and utterances were annotated using an entailment classifier to incorporate the entailment information while selecting responses. Finally, we follow a multi-layer keyword mining strategy to match meaningful concepts appearing in the context and response. 
\subsection{Emotion}
We trained an emotion classifier on \texttt{GoEmotions} dataset \cite{demszky2020goemotions}. This dataset contains  58k English Reddit comments, labeled for 27 emotion categories or Neutral. We fine-tuned \texttt{RoBERTa} using this dataset. We saved the checkpoint with the best Macro F1 of $49.4\%$ (equal to the current state of the art) and used this for annotating each utterance. Since emotion classification is a challenging task and given the complexity of the affect lexicons in the corpus, we only consider the labels which can be predicted with more than $90\%$ confidence(i.e., probability higher than $90\%$). The goal here is to study the effect of emotion in personalized response selection; developing a highly accurate emotion classifier is kept outside the scope of this work.
\subsection{Entailment}
For annotating entailment, we have used an ensemble of two models. The first one is \texttt{RoBERTa} based model trained on Stanford Natural Language Inference (SNLI) corpus \cite{snli-maccartney-manning-2008-modeling} released by AllenAI\footnote{\href{https://github.com/allenai/allennlp-models}{https://github.com/allenai/allennlp-models}}. The second model is also a \texttt{RoBERTa} based model fine-tuned on DECODE \cite{nie2020i}. We take the two models' weighted average of both probabilities during inference. The second model has a higher preference with $80\%$ weightage as it is trained on conversational data. The entailment label is assigned to every persona-response and utterance-response pair.
\subsection{Concept Mining} \label{concept-min}
We mine keywords and key phrases from the persona sentences, utterances, and responses denoted as $\{pc_i\}_{i=1}^{N_{pc}}, \{uc_i\}_{i=1}^{N_{uc}}, \{rc_i\}_{i=1}^{N_{rc}}$ respectively. We follow the techniques proposed in \cite{tang2019targetguided} to extract the first level of keywords. Subsequently, we expand the concept lists by extracting key phrases using the RAKE \cite{rake}. We hypothesize that concepts appearing in responses should adhere to the speaker's persona. So, we prune some of the response/ context keywords by calculating the average of Point-wise Mutual Information score between persona keywords and response/ context keywords $\sum_{j=1}^{N_{pc}} PMI(pc_j, rc_i)/N_{pc}$ and rejecting the concepts which are below a threshold value($\lambda$). Similarly, for response/ concept key phrases extracted using RAKE, we only keep top $N$ key phrases. Finally, we combine the persona and context keywords and treat them as context keywords($uc_i$).
\section{Methodology}
\subsection{Problem Definition}
Given a dataset $D = \left \{ (C_i, uc_i, p_i, r_i, rc_i, y_i) \right \}_{i=1}^{N} $ is a set of $N$ tuples consisting context $C_i$,  the persona of the speaker or the partner $p_i$, response to the context $r_i$, and the ground truth $y_i$. A set of concepts appearing in context and a response is denoted by $uc_i$ and $rc_i$, respectively. The context can be represented as $C_i = \left \{ (U_j, Emo_j, Entail_j) \right \}_{j=1}^{L}$ where $U_j$ is an utterance, $Emo_j$ is the dominant emotion present in $U_j$ and $Entail_j$ is the entailment label of  $U_j$ with respect to $r_i$. The $j^{th}$ utterance $U_j$ is denoted by $U_j = \left \{ u_1j, u_2j,..., u_Mj \right \}$ which consists of $M$ tokens. Each response $r_i$ contains single utterance,  $y_i \in \left \{ 0, 1 \right \}$, $Emo_j \in \left \{ 0, 1, ... P \right \}$ , and  $Entail_j \in $ \scalebox{0.85}{\{ \texttt{entailment}, \\ \texttt{neutral}, \\ \texttt{contradiction}\}} where $P$ are the total number of emotion types possible in the $D$. The task is to train a matching model for $D$, $g(C,uc,p,rc,r)$. Given a triple of context-persona-response the goal of the matching model $g(C,uc,p,rc,r)$ is to calculate the degree of match between $(C,uc,p)$ and $(rc,r)$.
\subsection{Bot Context Encoding} \label{bot-context}
When two users communicate, many topics are often discussed in parallel, and sometimes a few utterances might not be relevant for response selection. To account for the model to be aware of the speaker change information, \citeauthor{bertsa} introduced a speaker disentanglement strategy in the form of \textit{speaker embeddings} fused with the original token embeddings. This technique has proven to improve response selection performance \cite{bertsa, hclbert}. However, the problem of the maximum length of positional embeddings still exists. To circumvent this, we have created bot-context encoding, which captures the representation of the bot's turns in the context while ignoring the user's turns. The intent is to use the bot's turns to maintain consistency during response selection. The input sequence that is sent to BERT to encode bot context is composed as follows:

{
\small
\begin{eqnarray}
        x_{si} &=& [CLS] u_{2} [EOU] u_{4} [EOU]... \nonumber \\
                 & &... u_{n-1}[EOU][SEP]r_i[EOU]
\end{eqnarray}
}
Where ${u_{1}, u_{4}, ... u_{n-1}}$ are bot's utterances in the context, $[EOU]$ is a special token denoting the end of an utterance.

The resultant tokens $x_{si}$ are passed through \texttt{bert-base-uncased}, the last hidden states of $k$ layers i.e. $\{ \mathbf{h_{s1}^{(l)}, h_{s2}^{(l)},.. h_{sT}^{(l)}}\}$, for $l=1,2, ..k $ are used in downstream tasks.
\subsection{Fusion Strategies}
We use several fusion strategies to model the inter-dependencies of the persona, emotion, and entailment information. We use BERT \cite{devlin2019bert} as our base sentence encoder. Like the Bi-encoder \cite{humeau2020polyencoders} we concatenate context utterances as a single context sentence before passing it into BERT.
\subsubsection{Baseline} \label{baseline}
For the baseline, we have extended \texttt{BERT-CRA} \cite{partner} where persona and context are concatenated to form sequence A and response form sequence B. Then, these two sequences are concatenated using $[SEP]$ token. We made two changes to this model; first, we added \textit{speaker embeddings} with the original token representation. Secondly, we fuse bot-context encoding as described in the previous section with \texttt{BERT-CRA} encoding by doing multi-headed attention between the hidden representation of the last $k$ layers of both encoders. The token arrangement is as follows:

{
\small
\begin{eqnarray}
        x_{CRAi} &=& [CLS] p_1 p_2 ... p_i [EOP] u_{1} [EOU] \nonumber \\
                 & &... u_{i}[EOU][SEP]r_i[EOU] 
\end{eqnarray}
}
\begin{figure*}[t]
     \centering
     \begin{subfigure}{\textwidth}
        \centering
        \includegraphics[width=\textwidth]{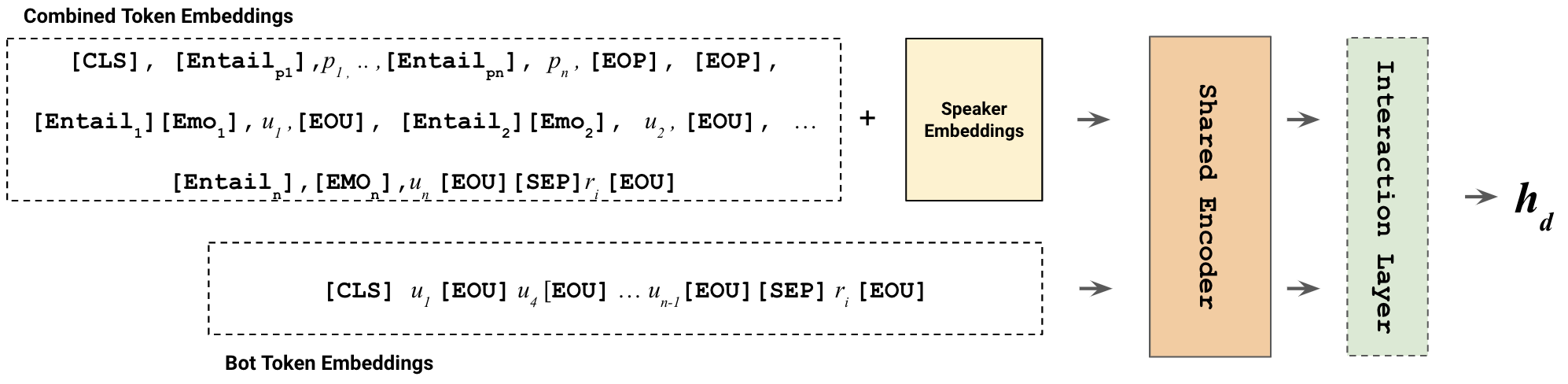}
        \caption{Dual encoder pipeline consisting of combination of all the encoding strategies.}
        \label{fig:combined_encoding}
     \end{subfigure}
     
     \begin{subfigure}[b]{0.8\textwidth}
         \centering
         \includegraphics[width=\textwidth]{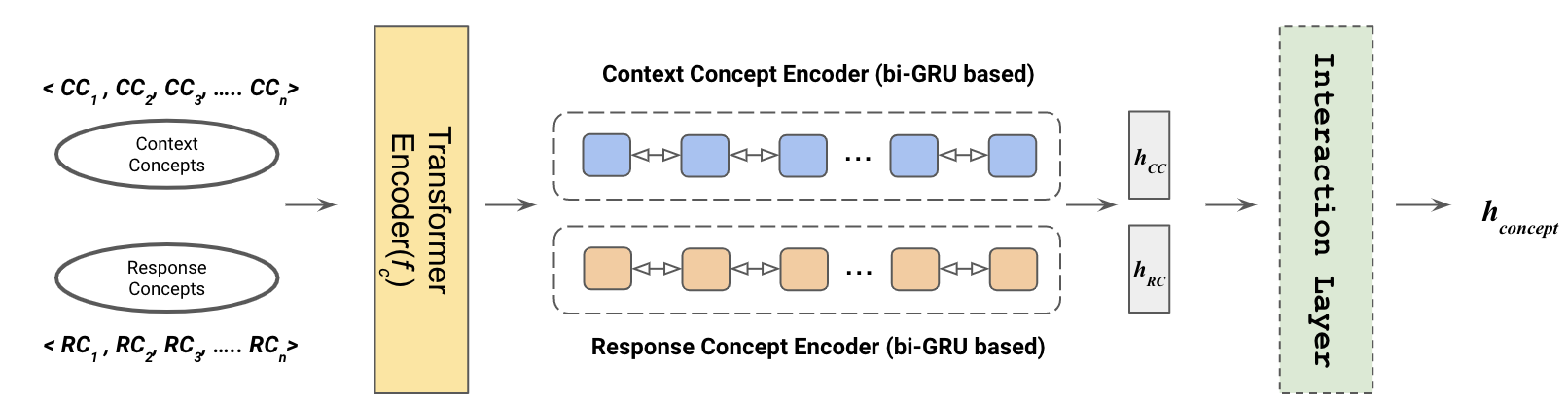}
         \caption{Concept-flow interaction network, the output of this network $\mathbf{h_{concept}}$ can be concatenated with any of the \texttt{BERT} based dual encoder's output($\mathbf{h_d}$).}
         \label{fig:concept_flow}
     \end{subfigure}
        \caption{Overall Training Architecture. Though the \texttt{BERT} based dual encoder is independently depicted but it is trained along with concept-flow interaction(if included).}
        \label{fig:three graphs}
        
\end{figure*}
Where $p_1 p_2 ... p_i$ are the personalities of the speaker, $[EOP]$ token denotes end of personality representation, $u_1, u_2,.. u_i$ are the utterances in the context. The resultant tokens $x_{CRAi}$ are passed through \texttt{bert-base-uncased}, the hidden states of last $k$ layers i.e. $\{ \mathbf{h_{c1}^{(l)}, h_{c2}^{(l)},.. h_{cT}^{(l)}}\}$, for $l=1,2, ..k $ are used in downstream tasks.

\textbf{Interaction Layer : } Since we are using a multi-encoder pipeline, it is crucial to capture the interaction between the encoders. For that, we use multi-head attention between hidden states of speaker context encoder and \texttt{BERT-CRA}. For ease of presentation, we denote the whole multi-headed attention layer as $f_{mha}(*, *)$. Then these attention outputs are passed through an aggregation layer, which basically concatenates then passes it through a two-layer feed-forward network and finally mean pools across all the layers to get $\mathbf{h_d}$. The output is passed through a $MLP$ to get the matching degree with the response. 

{
\small
\begin{eqnarray}
    \{ \mathbf{\widetilde{h}_{s1}^{(l)}, \widetilde{h}_{s2}^{(l)},.. \widetilde{h}_{sT}^{(l)}}\} &=& f_{mha}(\{ \mathbf{h_{s1}^{(l)}, h_{s2}^{(l)},.. h_{sT}^{(l)}}\} ,\nonumber \\
                  & &\{ \mathbf{h_{c1}^{(l)}, h_{c2}^{(l)},.. h_{cT}^{(l)}}\}) \\
    \{ \mathbf{\widetilde{h}_{c1}^{(l)}, \widetilde{h}_{c2}^{(l)},.. \widetilde{h}_{cT}^{(l)}}\} &=& f_{mha}(\{ \mathbf{h_{c1}^{(l)}, h_{c2}^{(l)},.. h_{cT}^{(l)}}\} ,\nonumber \\
                  & &\{ \mathbf{h_{s1}^{(l)}, h_{s2}^{(l)},.. h_{sT}^{(l)}}\}) \\
    \mathbf{h_d} &=& \mathrm{MeanPool}(\{\mathrm{FFN}([\{ \mathbf{\widetilde{h}_{s1}^{(l)},..\widetilde{h}_{sT}^{(l)}}\}; \nonumber \\
                 & & \{ \mathbf{\widetilde{h}_{c1}^{(l)},..\widetilde{h}_{cT}^{(l)}}\}])\}_{l=0}^k)
\end{eqnarray}
}
\begin{figure}[h]
\centering
\includegraphics[width=0.5\textwidth]{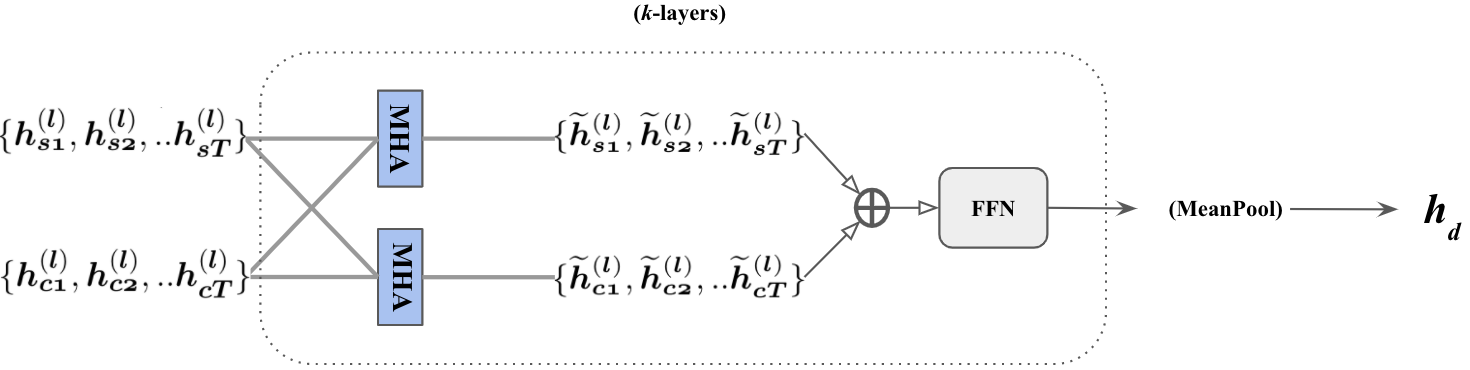}
\caption{Interaction Layer }
\label{fig:sp_encoding}
\end{figure}

\textbf{Loss Function: } The MLP layer predicts whether a context-persona $(C,p)$ pair matches with the corresponding response $r$ based on the derived features. Subsequently, the output from the MLP layer is passed through a softmax output layer to return a probability distribution over all response candidates. All the models described in this paper are trained using MLP cross-entropy loss. Let $\Theta $ be the model parameters, then the loss function $\mathcal{L}(D, \Theta)$ for all the models can be formulated as follows:

{
\small
\begin{equation}
    \mathcal{L}(D, \Theta) = - \sum_{(C,p,r,y))\epsilon D}^{} ylog(g(C,p,r))
\end{equation}
}
     

\subsubsection{ \texttt{BERT-EmA} Emotion Aware Fusion: }In this strategy, an emotion incorporation framework is introduced. Similar to \texttt{BERT-CRA} a dual pipeline matching network is followed. The first pipeline encodes the speaker's and listener's emotional and personality characteristics in the context. While the other encodes the bot-context as described in section \ref{bot-context}. 

We attach the most probable emotion tag to each utterance to incorporate emotion features in the BERT contextual representation. The emotion-infused context representation is then concatenated with the original persona representation, as described in section \ref{baseline}. The main goal of representing the context in this way is to understand how the emotions of each utterance interact with the speaker's persona. The input to the emotion encoder is as follows:

{
\small
\begin{eqnarray}
    x_{EmAi} &=& [CLS] p_1 p_2 ... p_i [EOP][Emo_1] u_{1} [EOU] \nonumber \\
             & &... [Emo_{i}]u_{i}[EOU] [SEP]r_i[EOU]
\end{eqnarray}
}
Similar to baseline, the hidden states of last  $k$ layers i.e. $\{ \mathbf{h_{e1}^{(l)}, h_{e2}^{(l)},.. h_{eT}^{(l)}}\}$, for $l=1,2, ..k $ are used in downstream tasks.
\subsubsection{ \texttt{BERT-EnA-P}: Entailment Aware Fusion} In this fusion strategy, the intention is to model the entailment information about each of the utterances and personas with the response. Like \texttt{BERT-EmA} we follow a dual encoder pipeline, the first encodes the entailment features, and the second encodes the bot-context.
To incorporate entailment features into BERT contextual representation, we attach entailment tags i.e. \texttt{<contradiction>}, \texttt{<entailment>}  and  \texttt{<neutral>} at the start of every utterance and persona. The response is concatenated with the context-entailment representation with a $[SEP]$ token. The input to the entailment encoder is as follows:

{
\small
\begin{eqnarray}
    x_{EmA-Pi} &=& [CLS][Entail_{p1}]p_{1} ... [EOP] \nonumber \\
             & & [Entail_{1}]u_{1} [EOU] \nonumber \\
             & & [Entail_{2}]u_{2} [EOU] \nonumber \\
             & &... [Entail_{i}]u_{i}[EOU] \nonumber \\
             & &    [SEP]r_i[EOU] 
\end{eqnarray}
}
The hidden states of last $k$ layers i.e. $\{ \mathbf{h_{en1}^{(l)}, h_{en2}^{(l)},.. h_{enT}^{(l)}}\}$, for $l=1,2, ..k $ are used in downstream tasks.

\noindent Finally, we experiment with a combined pipeline as depicted in Figure \ref{fig:combined_encoding}.


\subsection{Concept-Flow(CF) Interaction}


In section \ref{concept-min}, we describe how we extract relevant concepts from the context and the response. An appropriate response often has concepts most recently discussed in the context. So, to model that, we construct a concept-flow interaction network, where the interaction between the context-concepts and response-concepts are measured and used as a feature in response relevance classification.  

Let us consider $\left \{ {CC_1, CC_2, ..., CC_n} \right \}$ are concepts extracted from context and $\left \{ {RC_1, RC_2, ..., RC_n} \right \}$ are concepts extracted from a response. Now, we pass each of these concepts through a transformer-based concept encoder $f_c$ to get two sets of concept embeddings $\left \{ \mathbf{ec_1, ec_2, ..., ec_n} \right \}$ , $\mathbf{ec_i} \in  \mathbb R^{d_c} $  and $\left \{ \mathbf{rc_1, rc_2, ..., rc_n} \right \}$ , $\mathbf{rc_i} \in  \mathbb R^{d_c} $ for context and response concepts respectively. To learn the context flow representation for each set of concepts, we apply a bi-directional GRU network to capture sequential dependencies between subsequent concepts in a conversational situation. Context-concept and response-concept representation $\mathbf{h_i^{cc}}$ , $\mathbf{h_i^{rc}}$ can be formulated as:

{
\small
\begin{eqnarray}
    \mathbf{c_i^{cc}, h_i^{cc}} & = & \overleftrightarrow{GRU}(\mathbf{ec_i, h_{i-1}^{cc}}) \\
    \mathbf{c_i^{rc}, h_i^{rc}} & = & \overleftrightarrow{GRU}(\mathbf{er_i, h_{i-1}^{rc}}) \\
    \mathbf{h_{cc}} & = & tanh(\mathbf{\sum_{j \in 2*N_{l}} W_jh_j^{cc}}) \\
    \mathbf{h_{rc}} & = & tanh(\mathbf{\sum_{j \in 2*N_{l}} W_jh_j^{rc}})
\end{eqnarray}
}

Where $\mathbf{h_i^{cc}} \in  \mathbb R^{2d_c}$ , $\mathbf{h_i^{rc}} \in  \mathbb R^{2d_c}$ are the $i$ - the hidden  states and $\mathbf{c_i^{cc}} \in  \mathbb R^{2d_c}$ , $\mathbf{c_i^{rc}} \in  \mathbb R^{2d_c}$ are the outputs of the respective GRU encoders,  $\mathbf{W_j}$ is a learn-able parameter and $N_l$ is the number of layers in each GRUs. To model the interaction between $\mathbf{h_i^{cc}} $ and $\mathbf{h_i^{rc}}$ we follow the same interaction mechanism described in the earlier section. The output $\mathbf{h_{concept}}$ is concatenated with the dual encoder output $\mathbf{h_d}$ before passing it through an MLP.


\begin{table*}[t]
\centering
\resizebox{12cm}{!}{%
\begin{tabular}{@{}l|cccc|cccc@{}}
\toprule
\multirow{3}{*}{Model} & \multicolumn{4}{c|}{Self Persona}                               & \multicolumn{4}{c}{Partner Persona}                            \\ \cmidrule(l){2-9} 
                       & \multicolumn{2}{c|}{Original}    & \multicolumn{2}{c|}{Revised} & \multicolumn{2}{c|}{Original}    & \multicolumn{2}{c}{Revised} \\ \cmidrule(l){2-9} 
 &
  \textbf{hits@1} &
  \multicolumn{1}{c|}{\textbf{MRR}} &
  \textbf{hits@1} &
  \textbf{MRR} &
  \textbf{hits@1} &
  \multicolumn{1}{c|}{\textbf{MRR}} &
  \textbf{hits@1} &
  \textbf{MRR} \\ \midrule
FT-PC \cite{mazare-etal-2018-training}                & -    & \multicolumn{1}{c|}{-}    & 60.7          & -            & -    & \multicolumn{1}{c|}{-}    & -            & -            \\
DIM  \cite{gu-etal-2019-dually}                  & 78.8 & \multicolumn{1}{c|}{86.7} & 70.7          & 81.2         & 64.0 & \multicolumn{1}{c|}{76.1} & 63.9         & 76.0         \\
TransferTransfo \cite{wolf2019transfertransfo}       & 80.7 & \multicolumn{1}{c|}{-}    & -             & -            & -    & \multicolumn{1}{c|}{-}    & -            & -            \\
FIRE \cite{gu-etal-2020-filtering}               & 81.6 & \multicolumn{1}{c|}{-}    & 74.8          & -            & -    & \multicolumn{1}{c|}{-}    & -            & -            \\
BERT-CRA \cite{partner}              & 84.3 & \multicolumn{1}{c|}{90.3} & 79.4          & 86.9         & 71.2 & \multicolumn{1}{c|}{80.9} & 71.8         & 81.5         \\ \midrule
Baseline               & 84.4 & \multicolumn{1}{c|}{90.7} & 79.4          & 87.6         & 71.2 & \multicolumn{1}{c|}{81.1} & 71.4         & 81.5         \\
BERT-EmA               & 84.6 & \multicolumn{1}{c|}{90.9} & 79.8          & 87.7         & 71.4 & \multicolumn{1}{c|}{81.2} & 71.4         & 81.6         \\
BERT-P-EnA             & 85.3 & \multicolumn{1}{c|}{91.2} & 80.5          & 87.9         & 71.7 & \multicolumn{1}{c|}{81.3} & 71.3         & 81.4         \\
BERT-EmA+BERT-P-EnA    & 85.8 & \multicolumn{1}{c|}{91.4} & 80.7          & 88.0         & 72.3 & \multicolumn{1}{c|}{81.5} & 71.7         & 81.5         \\
BERT-EmA+BERT-P-EnA+CF (All) &
  \textbf{86.6*} &
  \multicolumn{1}{c|}{\textbf{91.6*}} &
  \textbf{81.3*} &
  \textbf{88.6*} &
  \textbf{72.6*} &
  \multicolumn{1}{c|}{\textbf{81.9*}} &
  \textbf{72.4*} &
  \textbf{81.9*} \\ \bottomrule
\end{tabular}%
}
\caption{Performance of the proposed and previous methods on the Persona-Chat dataset under different persona configurations. The meanings of "Self Persona", "Partner Persona", "Original", and "Revised" can be found in Section \ref{dataset}. "-" represents that the results were not reported in their papers. Numbers marked with * denote that the improvement over the best performing baseline is statistically significant (t-test with p-value < 0.05). Numbers in bold denote the combined fusion strategy that achieves the best performance. }
\label{tab:main_results}
\end{table*}

\section{Experimental Setup}
\subsection{Training Details}
The ratio of positive to negative samples in the training set is 1:19, so there is a high imbalance in training data. Taking inspiration from \cite{partner} we adopted a dynamic negative sampling strategy in which the ratio of positive and negative responses is 1:1 in an epoch. We keep the positive response constant and change the negative response for every epoch, generating data for 19 epochs. We use \texttt{bert-base-uncased} as the base for each of our pretraining-based fusion models. In concept mining strategy, we have taken the top 3 concepts extracted using RAKE, $\lambda$ for PMI-based scoring was varied from 0.3 to 0.8 with 0.1 steps, and 0.5 was found optimum. The number of turns in the conversation history used for concept mining varied following this set: $\{2,3,4,5,6,7\}$. We preserve the original parameters of \texttt{bert-base-uncased}. The number of $k$-last layers in the interaction layer varied following this set: $\{3,4,5,6\}$; after some initial experimentation, 4 was found as the optimum value. The number of heads in the multi-head attention layer was kept 8. We use a 6-layered version MiniLM\cite{wang2020minilm} to encode the concepts; the embedding dimension was 384. The number of layers in the bi-directional GRUs in the concept encoder is 2. A dropout rate of 0.7 is applied to the concept encoder hidden representation before we send it to the interaction layer. AdamW\cite{loshchilov2019decoupled} optimizer was used for optimization. The initial learning rate was set to 2e-5 and linearly decayed by L2 weight decay. The maximum sequence length was set to 320. The training batch size was 12. The relevance prediction head used a single feed-forward layer with sigmoid activation. All code was implemented using the PyTorch framework. Also, we used 2 NVIDIA RTX A5000 GPUs to train the models. The average training time for one epoch was 46 minutes, using all our fusion strategies and concept encoding.
\subsection{Evaluation Metrics}
We used the same evaluation metrics as the previous work to ensure comparable results. Each model aimed to select
the best-matched response from available candidates for the given context and persona. We calculated the recall of the true positive replies, denoted as \textbf{hits@1}. In addition, the mean reciprocal rank \textbf{(MRR)} was also adopted to take the rank of the correct response overall candidates into consideration.
\subsection{Comparison Methods}
For comparison, we have only selected pretraining-based models.
\begin{itemize}
\itemsep-0.3em 
    \item \textbf{FT-PC \cite{mazare-etal-2018-training}:} employed the “pretrain and fine-tune” framework by first pretraining on a domain-specific corpus, dialogues of which were extracted from Reddit, and then fine-tuning on the Persona-Chat.
    
    \item \textbf{DIM \cite{gu-etal-2019-dually}:} used a dually matching network (DIM) which performs interactive matching between responses and contexts and between responses and personas respectively for ranking response candidates.

    \item \textbf{TransferTransfo \cite{wolf2019transfertransfo}}: the paper fine-tunes a transformer model(GPT) using \texttt{Persona-Chat} dataset on a multi-task objective which combines several unsupervised task.
    
    \item \textbf{\texttt{BERT-CRA} \cite{partner}:} This work presents four context-aware persona fusion strategies and the models are initialized and pretrained using BERT on \texttt{Persona-Chat} dataset.
\end{itemize}
\subsection{Experimental Results}
Table \ref{tab:main_results} reports the evaluation results of our proposed and previous methods on \texttt{Persona-Chat}  under various persona configurations.  We can see that incorporating the emotion and entailment knowledge of the utterances coupled with generic distributional semantics and external knowledge learned from pretraining rendered improvements on both \textbf{hits@1} and \textbf{MRR} conditioned on various personas. Compared to FT-PC \cite{mazare-etal-2018-training}, our best model outperformed it by 20.4 \% in terms of hits@1 conditioned emotion, entailment and concepts. Compared to TransferTransfo \cite{wolf2019transfertransfo}, which was also trained using pretrained transformer models, our combined model outperformed it, which shows the effectiveness of fusion strategies and the concept-encoder. Lastly, our combined model outperformed the \texttt{BERT-CRA} \cite{partner} in all the tasks. We see a 2.3 \% and 1.9 \% improvement in original and revised self-persona, and 1.4 \% and 0.6 \% improvement in original and revised partner-persona in terms of \textbf{hits@1}. The results bolster our hypothesis that emotion, entailment, and concepts play an important role in the task of response selection. Also, it is to be noted that \texttt{Persona-Chat} is a synthetic dataset, i.e., the data collection did not happen naturally. Therefore, the chances that the user will display this subtle interplay of persona and emotion is less. In addition to that, we observe the presence of contradictory distractor responses. We see a significant performance improvement from this information by introducing entailment-aware fusion and concept encoding.
\subsection{Human Evaluation}
\begin{figure}[h]
\centering
\includegraphics[width=0.3\textwidth]{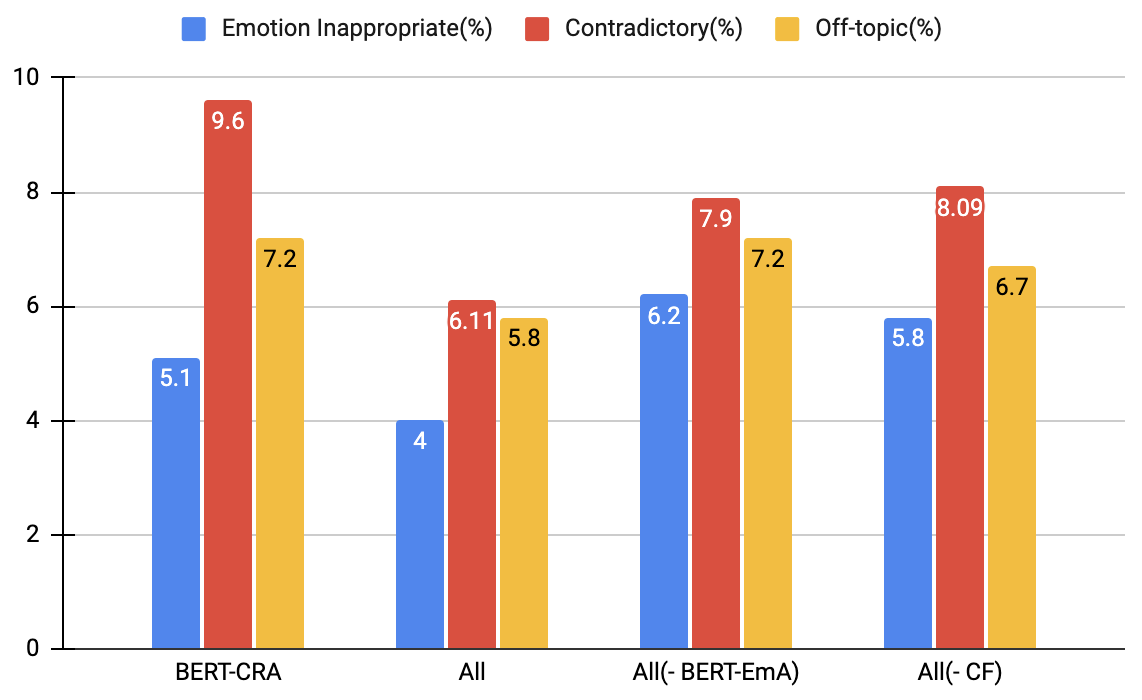}
\caption{Human evaluation results on \texttt{Persona-Chat} self-persona original test split}
\label{fig:human}
\end{figure}
Since a qualitative study by humans is necessary to understand the effectiveness of the proposed methods, we further perform a human evaluation on a portion of the data. We randomly sampled 100 inferred examples from the test set by baseline model, combined model, combined model except for the emotion, and combined model except for the concept flow interaction. We combined all the samples and evaluated them using Amazon Mechanical Turk by 2 different turkers on three metrics: emotion inappropriate, contradictory, and off-topic. The turkers needed to select if any of the three issues were present in an example. The percentages of reported issues by both groups are shown in Figure \ref{fig:human}. The results reveal that all our encoding pipelines effectively reduce contradictory responses and somewhat reduce off-topic and emotionally inappropriate responses. The agreement between the two groups was moderate (Krippendorff's $\alpha = 0.713$). 
\section{Analysis}
\subsection{Ablation Study for Emotion and Entailment}
We perform ablation studies(shown in Table \ref{tab:ablation}) to validate the effectiveness of emotion and entailment fusion in our proposed models. We see a slight improvement in our baseline model that uses our proposed speaker embedding. Also, unsurprising that the effect of emotion is not significant. As the dataset is artificially created, and emotions exhibited by the annotators are not always true. However, some performance improvement is observed. Conditioning persona in entailment fusion improves performance considerably as responses may not entail the speaker's persona.
\begin{table}[t]
\centering
\scalebox{0.7}{%
\begin{tabular}{@{}lcc@{}}
\toprule
\textbf{Models}             & \textbf{hits@1} & \textbf{MRR} \\ \midrule
Baseline                    & 84.4            & 90.7         \\
BERT-EmA($-$ Speaker Encoding) & 84.5            & 90.8         \\
BERT-EmA                    & 84.6            & 90.9         \\
BERT-EnA-P                  & \textbf{85.3}            & \textbf{91.2}         \\ \bottomrule
\end{tabular}}
\caption{ Ablation Study for Emotion and Entailment on self-original persona.}
\label{tab:ablation}
\end{table}
\subsection{Effect of Context Turns on Concept Representation}
\begin{figure}
\centering
\includegraphics[width=0.3\textwidth]{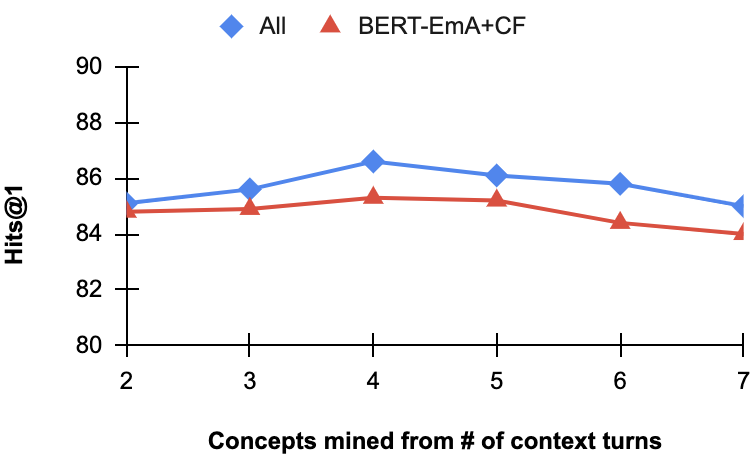}
\caption{This graph shows how \textbf{hit@1} reaches an optimum value and then decreases with an increase in the number of turns used to mine concepts.} \label{fig:M1}
\end{figure}

Concept matching boosts the evaluation performance further. However, the number of turns in the conversation history from which we mine the concepts influences the performances. It is evident from Figure \ref{fig:M1}  that the essential concepts in the most relevant response will be present in the recent conversation history.  
\subsection{Case Study}
\begin{table}[t]
\centering
\scalebox{0.5}{%
\begin{tabular}{@{}l|p{13cm}l@{}}
\toprule
personas &
  \begin{tabular}[c]{@{}p{11cm}l@{}}my favorite color is \colorbox{pink}{blue}. \textcolor{blue}{<ent: neutral>} \\ I enjoy \colorbox{pink}{reading mysteries}. \textcolor{blue}{<ent: neutral>}\\ I have \colorbox{pink}{seven children}. \textcolor{blue}{<ent: entail>}\\ I \colorbox{pink}{grew up} on a \colorbox{pink}{large farm}. \textcolor{blue}{<ent: neutral>}\end{tabular} \\ \midrule
context &
  \begin{tabular}[c]{@{}p{11cm}l@{}}A: hello how are you today? \textcolor{red}{<emo:curiosity>} \textcolor{blue}{<ent: neutral>}\\ B: I am well. how are you? \textcolor{red}{<emo:curiosity>} \textcolor{blue}{<ent: neutral>} \\ A: I am doing great just got back from the beach \textcolor{red}{<emo:excitement>} \textcolor{blue}{<ent: neutral>}\\ B: that is great. I live far from the \colorbox{pink}{beach}. \textcolor{red}{<emo:caring>} \textcolor{blue}{<ent: neutral>} \\ A: I am very lucky we live beside the beach. what do you do for a living? \textcolor{red}{<emo:curiosity>} \textcolor{blue}{<ent: neutral>} \\ B: I keep busy with my \colorbox{pink}{seven children}. \textcolor{red}{<emo:excitement>} \textcolor{blue}{<ent: neutral>} \\ A: wow that much have taken some adjusting I \colorbox{pink}{teach kindergarten}. \textcolor{red}{<emo:surprise>} \textcolor{blue}{<ent: neutral>} \end{tabular} \\ \midrule
\begin{tabular}[c]{@{}l@{}}golden\\ response\end{tabular} &
  do you teach mysteries to your children ? they are my favorite type of novel . \textcolor{red}{<emo:curiosity>} \\ \midrule
BERT-CRA &
  that must be a lot of \colorbox{pink}{work} but \colorbox{pink}{very rewarding} i bet  \textcolor{red}{<emo:realization>} \\ \midrule
\begin{tabular}[c]{@{}l@{}}All\end{tabular} &
  do you teach \colorbox{pink}{mysteries} to \colorbox{pink}{your children} ? they are my favorite type of \colorbox{pink}{novel} . \textcolor{red}{<emo:curiosity>}\\ \bottomrule
\end{tabular}}
\caption{ Case study showing concept flow.}
\label{tab:case_study}
\end{table}
Table \ref{tab:case_study} shows the efficacy of concept-encoding. Sometimes models fine-tuned on pretrained transformer models like BERT-CRA tends to select more generic responses. These models pay less attention to the persona or specific keywords in the context while selecting responses. In this example, our proposed model performs better than BERT-CRA as it is conditioned on the concepts. Specifically, concepts in the correct response i.e \ul{"mysteries"}, \ul{"novel"} relates to \ul{"reading mysteries"} concept in the persona and \ul{"your children"} relates to \ul{"teach kindergarten"} in the context. 
\section{Conclusion}
This work proposes a suite of novel fusion strategies and concept-flow encoder, which leverages the utterances' emotion, entailment, and concept information. These features help improve the performances of our models and provide critical insights into certain aspects of how humans communicate with each other. Though the techniques used in this paper are simple, it highlights the areas where response selection often falters, like detecting contraction, deviation from the concepts, etc. This work can be further extended by using a graphical model to improve the concept representations.

\section*{Acknowledgements}
We thank the anonymous reviewers for providing valuable feedback on our manuscript. This work is partly supported by NSF grant number IIS-2214070.  The content in this paper is solely the responsibility of the authors and does not necessarily represent the official views of the funding entity.

\bibliography{anthology,custom}
\bibliographystyle{acl_natbib}

\appendix

\section{Appendix}
\label{sec:appendix}

\subsection{Human Evaluation for Initial Study} \label{a1}

Five hundred context response pairs randomly sampled from the Persona-Chat self-original test split inferred by \texttt{BERT-CRA} were evaluated by at least two AMT workers. The following questions were asked to the workers:

\begin{enumerate}
    \item{Is this response contain emotions that are consistent with the context? (Most definitely/ not at all)}
    \item{Is this response contradicts the context? (Most definitely/ not at all)}
    \item{Do the topics discussed in this response appropriate to the topics discussed in the context? (Most definitely/ not at all)}
\end{enumerate}

\subsection{Final Human Evaluation}

Same evaluation pattern is followed as \ref{a1}, average pay for each HIT was 0.07 \$.

\end{document}